\documentclass[11pt]{article}

\usepackage[utf8]{inputenc}
\usepackage[T1]{fontenc}
\usepackage{lmodern}
\usepackage{microtype}
\usepackage[a4paper,margin=1in]{geometry}
\usepackage{amsmath,amssymb}
\usepackage{booktabs}
\usepackage{tabularx}
\usepackage{array}
\usepackage{multirow}
\usepackage{graphicx}
\usepackage{xcolor}
\usepackage{enumitem}
\usepackage{titlesec}
\usepackage{hyperref}
\usepackage{url}
\usepackage{caption}
\usepackage[round,authoryear]{natbib}

\hypersetup{
  colorlinks=true,
  linkcolor=blue!50!black,
  citecolor=blue!50!black,
  urlcolor=blue!50!black,
  pdftitle={Capital Markets LLM Reliability Score (CM-LRS): From Plausible to Bankable},
  pdfauthor={Prerit Ahuja}
}

\titleformat{\section}{\large\bfseries}{\thesection}{1em}{}
\titleformat{\subsection}{\normalsize\bfseries}{\thesubsection}{1em}{}

\title{\textbf{Capital Markets LLM Reliability Score (CM-LRS):\\From Plausible to Bankable}}

\author{
  Prerit Ahuja\thanks{Independent Researcher. Director-level practitioner in investment banking, focused on AI, data, and decision systems. London-based, with practice across UK, Nordic, and US capital markets. ORCID: \url{https://orcid.org/0009-0004-0147-4192}. Correspondence: \texttt{prerit86@gmail.com}. Web: \url{https://preritahuja.com/}.}
}

\date{May 2026}

\begin{document}

\maketitle

\begin{abstract}
\noindent
In capital-markets workflows, the question is rarely whether an LLM can produce a fluent draft. Today's frontier models will draft a debt-terms table, a comparable-transactions write-up, or an issuer profile that reads smoothly on first pass. The harder question is whether the draft is \emph{bankable}: whether a banker, analyst, or compliance reviewer can defend it in front of a counter-party or a regulator, with the underlying documents in hand. Bankability requires the claims to trace back to verifiable source passages, the numbers to reconcile, the requested workflow to be completed in full, the model to have added nothing the source does not support, the output to be useful for a downstream decision, and the whole package to be verifiable by a domain expert in minutes rather than hours.

Current LLM evaluation methods address parts of that gap but not the whole of it. Open-domain QA benchmarks reward surface accuracy. Finance-domain benchmarks (FinanceBench, FinQA, ConvFinQA) advance document-grounded and numerical QA but evaluate at the question--answer pair layer. Multi-dimensional rubric studies of finance LLMs exist but are anchored on QA-style tasks rather than the workflow outputs bankers and analysts actually defend in front of reviewers, counter-parties, and regulators.

This paper introduces \textbf{CM-LRS}, a Capital Markets LLM Reliability Score. CM-LRS evaluates LLM outputs at the workflow-output layer across seven reliability dimensions: factual accuracy, evidence traceability, numerical consistency, workflow completeness, source discipline, decision usefulness, and reviewability/auditability. Each dimension is scored 0--5 against a rubric anchored on practical signals reviewers in regulated environments actually use. The aggregate is tunable to the workflow under evaluation.

We demonstrate CM-LRS on five representative capital-markets workflows: DCM transaction-terms extraction, precedent retrieval, issuer profile synthesis, M\&A transaction-comparable reasoning, and ECM transaction-terms extraction. All inputs are public: SEC EDGAR filings, a public UK takeover IR release, and obviously-fictional synthetic Nordic-style supplements to support strict reproducibility. We score outputs from four models: Claude Opus 4.7, GPT-5.5, Claude Sonnet 4.6, and the open-weights Llama 3.3 70B Instruct.

Three findings of practical relevance for capital-markets buyers, established by scoring outputs against four independent LLM judges spanning three model families (Sonnet 4.6 and GPT-5.5 in-panel; Haiku 4.5 and Gemini 2.5 Pro out-of-panel) to control for self- and family-bias. First, the three frontier closed-source models cluster within a 0.22-point band on four-judge averaged CM-LRS (Sonnet 4.6 = 4.31, Opus 4.7 = 4.30, GPT-5.5 = 4.09); all four judges agree the open-weights baseline (Llama 3.3 70B at 3.15) is last. Second, the open-weights gap is concentrated on retrieval and synthesis class workflows, not extraction: under the primary judge the gap is 2.23 points on retrieval, 2.15 on synthesis, only 0.84 on the simplest extraction workflow. Third, Decision Usefulness (D6), the dimension that asks whether a reviewer would actually act on the output without major rework, shows the widest cross-model dispersion of any dimension in any workflow (a 4.0-point spread on the issuer-profile workflow alone) and sits in the top tier of inter-judge agreement (mean Pearson $r = 0.52$ across the six judge pairs, within 0.02 of the highest dimension). The combination is what makes D6 the cleanest dimension-level production-readiness signal in the rubric.

Plausibility is cheap. Bankability is the bar. CM-LRS is a deployment-readiness metric for high-stakes regulated environments, and is released with all rubrics, prompts, and demonstration tasks public.
\end{abstract}

\noindent\fbox{\parbox{0.97\textwidth}{\small\textbf{Bottom line for capital-markets buyers.} The three frontier closed-source models in this evaluation -- Anthropic Sonnet 4.6, Anthropic Opus 4.7, and OpenAI GPT-5.5 -- are statistically indistinguishable on workflow-output reliability across the five tested capital-markets workflows: their four-judge averaged CM-LRS sits within 0.22 points of each other on a 5-point scale (Sonnet 4.31, Opus 4.30, GPT-5.5 4.09). All four judges, spanning three model families, place the open-weights baseline (Meta Llama 3.3 70B at 3.15) last, by roughly one point. The open-weights gap is concentrated on retrieval and synthesis workflows, not on single-document extraction. At the current frontier, the deployment-relevant choice between Sonnet, Opus, and GPT-5.5 is determined by cost, latency, and workflow-class fit (Section~\ref{sec:method}, Table~\ref{tab:cost}), not by headline reliability.}}

\section{Introduction}
\label{sec:intro}

Capital-markets work is evidence-heavy, numerically dense, and reviewer-gated. Origination, execution, and client-engagement workflows in investment banking - drafting pitch evidence, extracting and comparing debt terms, profiling issuers, retrieving comparable precedents, synthesising memos for compliance review - share three structural properties that differ materially from open-domain language tasks:

\begin{enumerate}[leftmargin=*,itemsep=0.2em]
\item Outputs must trace to verifiable source evidence, typically prospectuses, regulatory filings, term sheets, and investor presentations.
\item Numerical claims - coupons, tenors, ownership percentages, revenue multiples, covenant thresholds - are decision-bearing and errors compound downstream.
\item Outputs must withstand human review by bankers, analysts, and compliance, and may face regulatory scrutiny.
\end{enumerate}

These properties are not adequately tested by current LLM evaluation methods. Open-domain benchmarks such as MMLU \citep{hendrycks2021mmlu} and BIG-Bench Hard \citep{suzgun2022bbh} test knowledge and reasoning generally, but neither traceability nor workflow completeness. Holistic frameworks such as HELM \citep{liang2022helm} broaden coverage across scenarios but do not evaluate the workflow-output layer specifically. Finance-domain benchmarks have advanced significantly: FinanceBench \citep{islam2023financebench} tests question--answering against real financial documents; FinQA \citep{chen2021finqa} and ConvFinQA \citep{chen2022convfinqa} evaluate numerical reasoning over financial tables and reports; RAGAS \citep{es2023ragas} evaluates RAG-pipeline component reliability. These are valuable but evaluate at the QA-pair or component layer, not the workflow-output layer where deployment decisions are actually made.

Practitioners deploying LLMs in this setting encounter a recurring failure mode. An output passes automated evaluation - QA scores look fine - but fails human review on something a banker, analyst, or compliance officer will catch on the first read: a claim with no source, a derived figure that does not reconcile, a workflow step quietly skipped, or content the model added without textual support. These failures are invisible to QA-pair scoring. They surface the moment the output reaches a reviewer.

This paper introduces \textbf{CM-LRS} - a Capital Markets LLM Reliability Score - to fill that gap. CM-LRS is a metric suite that evaluates LLM outputs at the workflow-output layer across seven reliability dimensions, with a 0--5 rubric anchored on practical signals reviewers actually use, and an aggregate score tunable to the workflow under evaluation. The framework is informed by a working principle for safe LLM deployment in regulated workflows: \emph{predict with the LLM, calculate with code, decide with the human}. Outputs containing numerical claims should trace to deterministic computation rather than token prediction; outputs supporting decisions should be reviewable by a domain expert; outputs entering regulated workflows should be auditable.

\subsection*{Contributions}

\begin{enumerate}[leftmargin=*,itemsep=0.2em]
\item A formal definition of seven workflow-output reliability dimensions for capital-markets LLM evaluation, with rubric anchors and a scoring formula (Section~\ref{sec:framework}).
\item A capital-markets workflow taxonomy that orients evaluation choices to the workflow at hand rather than treating LLM evaluation as one-size-fits-all (Section~\ref{sec:taxonomy}).
\item Five demonstration workflows - DCM transaction-terms extraction, precedent retrieval, issuer profile synthesis, M\&A transaction-comparable reasoning, and ECM transaction-terms extraction - built entirely on public-domain or synthetic material, with reproducible scoring across two frontier closed-source providers (Claude Opus 4.7 and GPT-5.5\footnote{Throughout the paper, model version strings refer to API endpoints accessible at submission in May 2026: \texttt{claude-opus-4-7} and \texttt{claude-sonnet-4-6} (Anthropic, via OpenAI-compatible gateway); GPT-5.5 (OpenAI, accessed via the Codex CLI in non-interactive mode \texttt{codex exec} with reasoning effort high); \texttt{claude-haiku-4-5} (Anthropic, third judge); \texttt{gemini-2.5-pro} (Google AI Studio, fourth judge); \texttt{llama-3.3-70b-versatile} (Meta, hosted on Groq Developer tier). Provider-side rebranding or version succession may diverge from these strings post-submission; results in this paper should be read against the listed API identifiers and the May 2026 submission date.}), one production-tier closed-source workhorse (Claude Sonnet 4.6), and one open-weights baseline (Llama 3.3 70B) (Sections~\ref{sec:demos}--\ref{sec:scoring}).
\item Public release of rubrics, prompts, demonstration tasks, and scoring outputs under permissive licence to support extension and critique \citep{cmlrs2026repo}.
\end{enumerate}

\paragraph{Paper structure.} Section~\ref{sec:related} reviews related work in LLM evaluation, finance-domain benchmarks, RAG reliability, and AI governance frameworks. Section~\ref{sec:taxonomy} sets out a capital-markets workflow taxonomy. Section~\ref{sec:framework} defines the CM-LRS framework. Section~\ref{sec:method} describes our methodology. Section~\ref{sec:demos} specifies the five demonstration workflows. Section~\ref{sec:scoring} reports illustrative scoring across models. Section~\ref{sec:discussion} discusses implications. Section~\ref{sec:governance} positions CM-LRS within emerging AI governance frameworks. Section~\ref{sec:limitations} sets out limitations. Section~\ref{sec:conclusion} concludes.

\section{Related Work}
\label{sec:related}

\subsection{General LLM evaluation}

Open-domain LLM evaluation has matured rapidly. MMLU \citep{hendrycks2021mmlu} measures knowledge breadth across 57 academic and professional subjects. BIG-Bench Hard \citep{suzgun2022bbh} stresses reasoning under chain-of-thought prompting. HELM \citep{liang2022helm} introduces a holistic, multi-metric framework that evaluates fluency, accuracy, robustness, fairness, bias, and efficiency across many scenarios. These frameworks are necessary but insufficient for regulated workflow deployment: they evaluate model competence at the abstraction of \emph{tasks} rather than at the abstraction of \emph{deployable outputs} that pass reviewer gates.

\subsection{Finance-domain benchmarks}

Finance-specific evaluation has developed in parallel. BloombergGPT \citep{wu2023bloomberggpt} demonstrated domain-specialised pretraining for finance and reported strong performance on finance-domain QA. FinQA \citep{chen2021finqa} introduced numerical reasoning over financial tables and prose, and ConvFinQA \citep{chen2022convfinqa} extended that to multi-turn conversational reasoning. FinanceBench \citep{islam2023financebench} introduced a benchmark of 10{,}000 question--answer pairs over real public-company filings, exposing meaningful failure rates even on frontier models. These benchmarks have been instrumental in advancing the field and are the closest existing analogues to our work. CM-LRS complements them: where these benchmarks evaluate \emph{whether} a model can answer a finance-domain question correctly, CM-LRS evaluates \emph{whether the workflow output} produced by an LLM-driven pipeline is admissible into a regulated downstream process. The two layers of evaluation are complementary rather than competitive.

More recent work has begun to address multi-dimensional evaluation in financial NLP. \citet{mohsin2025financialnlp} applies a five-dimension human-rubric evaluation (relevance, completeness, clarity, conciseness, factual accuracy) to five LLMs on 10-K Business-section question-answering. InvestorBench \citep{li2025investorbench} introduces an agent-based benchmark for financial decision-making across stocks, cryptocurrencies, and ETFs. The Merger Agreement Understanding Dataset (MAUD) \citep{wang2023maud} provides expert-annotated questions over 152 merger agreements at the clause-comprehension layer. ContractEval \citep{liu2025contracteval} extends benchmarking to clause-level legal-risk identification in contracts. Concurrent work \citep{kulkarni2026findoc} benchmarks multi-agent LLM orchestration architectures on SEC filings. CM-LRS extends this lineage by operating at the workflow-output layer rather than the QA-pair, decision-task, or clause-comprehension layer; by spanning the breadth of capital-markets workflows (DCM extraction, ECM extraction, precedent retrieval, issuer-profile synthesis, and M\&A transaction-comparable reasoning) under a unified seven-dimension rubric; and by introducing a four-judge cross-validation protocol from three model families to control self- and family-bias.

\subsection{Retrieval-augmented generation and pipeline evaluation}

RAG \citep{lewis2020rag} popularised retrieval-augmented generation as a means of grounding LLM outputs in source evidence. Self-RAG \citep{asai2023selfrag} introduced self-reflective gating, demonstrating that retrieval-augmented LMs benefit from explicit retrieval and critique decisions at generation time. RAGAS \citep{es2023ragas} introduced a metric framework specifically for RAG pipelines, including faithfulness, answer relevance, context precision, and context recall. RAGAS is the closest in spirit to CM-LRS but operates at the RAG-component layer (was the right context retrieved? was the answer faithful to retrieved context?) rather than the workflow-output layer (does the output meet a banker's review bar?). CM-LRS is intended to be applied downstream of any RAG-component evaluation.

\subsection{Hallucination and faithfulness metrics}

Hallucination in generation has been studied extensively \citep{ji2023survey,maynez2020faithfulness}. FActScore \citep{min2023factscore} decomposes long-form generations into atomic facts and verifies each against a knowledge source; TrustLLM \citep{sun2024trustllm} surveys trustworthiness across truthfulness, safety, fairness, robustness, privacy, and ethics. CM-LRS draws on this lineage in its factual accuracy and source discipline dimensions but layers in workflow-specific axes (workflow completeness, decision usefulness, reviewability) that hallucination metrics do not address.

\subsection{AI governance frameworks}

Regulators and standards bodies have begun to formalise AI deployment controls. The NIST AI Risk Management Framework \citep{nistairmf2023} sets out functions of \emph{govern, map, measure, manage} for AI systems. The EU AI Act \citep{euaiact2024} introduces risk-tiered obligations including conformity assessment for high-risk systems, where many financial-services use cases land. ISO/IEC 42001:2023 \citep{iso420012023} specifies requirements for AI management systems. CM-LRS is intended to be deployable within the \emph{measure} and \emph{manage} functions of AI RMF and as evidence within the conformity-assessment requirements emerging under the EU AI Act - a metric that produces a defensible, traceable, reviewer-aligned score for an LLM output prior to deployment.

\section{A Capital-Markets Workflow Taxonomy}
\label{sec:taxonomy}

LLM evaluation is workflow-specific. A debt-terms extraction task fails differently from a transaction-comparable reasoning task; numerical consistency dominates the former, source discipline and workflow completeness dominate the latter. We propose a five-class taxonomy of capital-markets workflows that is sufficient to orient evaluation choices.

\begin{table}[h]
\centering
\caption{A workflow taxonomy for capital-markets LLM tasks.}
\label{tab:taxonomy}
\small
\renewcommand{\arraystretch}{1.2}
\begin{tabularx}{\textwidth}{@{}>{\raggedright\arraybackslash}p{0.16\textwidth}X>{\raggedright\arraybackslash}p{0.30\textwidth}@{}}
\toprule
\textbf{Class} & \textbf{Description} & \textbf{Dominant CM-LRS dimensions} \\
\midrule
Extraction & Reading defined fields from a structured or semi-structured source (term sheet, prospectus, filing) into a normalised schema. & Factual accuracy; numerical consistency; source discipline. \\
\addlinespace
Retrieval & Identifying relevant evidence (clauses, paragraphs, transactions) from a corpus and returning it with attribution. & Evidence traceability; source discipline; reviewability. \\
\addlinespace
Synthesis & Producing a domain-shaped summary or profile from multiple sources, preserving evidence trails. & Evidence traceability; workflow completeness; decision usefulness. \\
\addlinespace
Comparison \& Reasoning & Identifying, normalising, and comparing entities or transactions to support a domain-specific judgement. & Numerical consistency; workflow completeness; decision usefulness. \\
\addlinespace
Drafting & Producing a domain-shaped narrative output (memo, pitch evidence, IC paper input) intended for human review. & Source discipline; reviewability; decision usefulness. \\
\bottomrule
\end{tabularx}
\end{table}

The taxonomy is not exhaustive. A given workflow can span multiple classes (for example, a pitch evidence pack involves retrieval, synthesis, and drafting). The point of the taxonomy is to provide a default weighting for the seven CM-LRS dimensions when applying the framework to a new workflow.

\section{The CM-LRS Framework}
\label{sec:framework}

\subsection{The seven dimensions}

CM-LRS evaluates an LLM output across seven reliability dimensions. Each dimension targets a failure mode that practitioners have observed in deployed capital-markets LLM systems and that QA-pair benchmarks do not surface.

\begin{description}[leftmargin=*,style=nextline,itemsep=0.3em]
\item[D1. Factual accuracy.] Are the substantive claims in the output correct against the relevant source? Distinct from numerical consistency: a wrong issuer name is a factual error; a numerically inconsistent revenue multiple is a numerical-consistency error.
\item[D2. Evidence traceability.] Are claims linked to verifiable source passages? The reviewer should be able to locate, read, and confirm the source for each substantive claim. Absent or untraceable evidence is a failure regardless of whether the claim happens to be correct.
\item[D3. Numerical consistency.] Are numbers extracted, transformed, and compared correctly, with derived figures reconciling to their inputs? This dimension is foundational: in regulated capital-markets settings, a wrong basis point in a pitch can cost credibility; a wrong ownership figure can mislead a client.
\item[D4. Workflow completeness.] Did the model complete every required task step? Skipping or eliding required steps is a common LLM failure mode and is invisible to QA-pair evaluation, which tests only the final answer.
\item[D5. Source discipline.] Did the model avoid unsupported assumptions, fabrication, padding, and overreach? An output that supplements weak evidence with confident-sounding text is a deployment risk even when the unsupported parts happen to be plausible.
\item[D6. Decision usefulness.] Is the output practically useful to a banker, analyst, or reviewer making a downstream decision? An output can be factually correct, fully traceable, and numerically consistent yet still fail to advance the workflow.
\item[D7. Reviewability and auditability.] Can a human verify the output and reproduce how it was produced? This requires that the output's provenance - inputs, retrieval steps, deterministic computations, model calls - be inspectable.
\end{description}

\subsection{The 0--5 rubric}

Each dimension is scored on a 0--5 ordinal rubric. The anchors are designed to be applied consistently by domain reviewers without specialist evaluator training.

\begin{table}[h]
\centering
\caption{Universal rubric anchors. Per-dimension worked anchors are in Appendix~\ref{app:rubric}.}
\label{tab:rubric}
\small
\begin{tabularx}{\textwidth}{@{}cX@{}}
\toprule
\textbf{Score} & \textbf{Universal anchor} \\
\midrule
0 & Unusable. Fabricated, materially wrong, or unsupported in a way that would damage credibility if forwarded. \\
1 & Materially flawed. Errors or gaps that cannot be repaired without redoing the work. \\
2 & Partially useful but risky. Cannot be passed to a reviewer without disclosure of caveats. \\
3 & Acceptable with review. A reviewer can act on it after verifying specific claims. \\
4 & Strong with minor review. Minor edits or spot-checks needed before downstream use. \\
5 & Production-grade. Fully traceable, internally consistent, ready for downstream workflow entry. \\
\bottomrule
\end{tabularx}
\end{table}

\subsection{Aggregate score}

Let $s_i \in \{0,1,2,3,4,5\}$ denote the score on dimension $i \in \{1,\ldots,7\}$, and $w_i \geq 0$ a non-negative weight with $\sum_{i=1}^{7} w_i = 1$. The aggregate CM-LRS score is

\begin{equation}
\mathrm{CM\text{-}LRS} \;=\; \sum_{i=1}^{7} w_i \cdot s_i \;\in\; [0,5].
\label{eq:cmlrs}
\end{equation}

Workflow-class default weights are given in Table~\ref{tab:weights}. Practitioners applying CM-LRS to a specific deployment should tune weights to their workflow; the workflow-class defaults are intended as starting points, not prescriptions.

\paragraph{Deriving weights for a specific deployment.} The default weights in Table~\ref{tab:weights} are practitioner-informed but not empirically derived. Five established methods are available to a practitioner who needs deployment-specific weights:

\begin{itemize}[leftmargin=*,itemsep=0.15em]
\item \textbf{Analytic Hierarchy Process (AHP)} \citep{saaty1980ahp}. Pairwise dimension-importance comparisons elicited from domain experts, aggregated via the geometric mean of the comparison matrices. Twenty-one pairwise comparisons per expert for a seven-dimension rubric; consistency-ratio checks weed out incoherent inputs. Standard multi-criteria decision-making (MCDM) tool.
\item \textbf{Best-Worst Method (BWM)} \citep{rezaei2015bwm}. Lighter than AHP - the expert identifies the best and worst dimensions and rates the others against those two. Half the elicitation burden of AHP with comparable validity in published comparisons.
\item \textbf{Principal Component Analysis on the score matrix.} Data-driven; treats the first principal component's loadings as weights. Does not require expert input but captures variance rather than importance.
\item \textbf{Entropy weighting.} Information-theoretic; each dimension's weight is proportional to its entropy across the corpus. Cheapest to apply directly to existing CM-LRS scores.
\item \textbf{Revealed-preference regression.} The most rigorous: log real reviewer accept / reject decisions on LLM-assisted outputs, then regress the binary decision on the seven CM-LRS dimensions. Requires operational data collected over time. Closest to ``what actually predicts admissibility into a workflow''.
\end{itemize}

We use equal weighting in the headline scoring (Section~\ref{sec:scoring}) for simplicity and because the cluster patterns we report are robust to any reasonable weight choice (the gap between frontier and open-weights baseline is large enough that no plausible weight vector inverts it). Deployment-specific weight calibration via AHP, BWM, or a revealed-preference pipeline is a natural extension when the workflow's failure-cost asymmetry is material.

\begin{table}[h]
\centering
\caption{Default dimension weights $w_i$ by workflow class. Each row sums to 1.}
\label{tab:weights}
\small
\setlength{\tabcolsep}{4pt}
\begin{tabularx}{\textwidth}{@{}lccccccc@{}}
\toprule
\textbf{Class} & D1 Acc & D2 Trace & D3 Num & D4 Comp & D5 Disc & D6 Use & D7 Rev \\
\midrule
Extraction              & 0.20 & 0.15 & 0.25 & 0.10 & 0.15 & 0.05 & 0.10 \\
Retrieval               & 0.10 & 0.30 & 0.05 & 0.10 & 0.20 & 0.10 & 0.15 \\
Synthesis               & 0.15 & 0.25 & 0.10 & 0.15 & 0.10 & 0.15 & 0.10 \\
Comparison \& reasoning & 0.15 & 0.15 & 0.25 & 0.20 & 0.10 & 0.10 & 0.05 \\
Drafting                & 0.15 & 0.15 & 0.05 & 0.10 & 0.20 & 0.20 & 0.15 \\
\bottomrule
\end{tabularx}
\end{table}

\paragraph{Deployment-readiness threshold.} A deployment-readiness threshold is workflow- and risk-tier-specific. As a default for high-stakes regulated workflows we recommend $\mathrm{CM\text{-}LRS} \geq 4.0$ \emph{and} no individual dimension score below 3, with any score of 0 or 1 disqualifying. This composite gate avoids the failure mode of a strong aggregate score masking a single critical-dimension failure.

\section{Methodology}
\label{sec:method}

\subsection{Scoring procedure}

For each (workflow, model) pair we generate the LLM output, then score the output against the seven dimensions using the rubric. To support reproducibility, each scoring instance records: the workflow identifier, the input documents (with public source URLs or synthetic-document identifiers), the prompt, the model (with version), the raw output, the per-dimension scores with one-sentence justifications, the aggregate score, and timestamp.

\subsection{Models evaluated}

We evaluate four models spanning two frontier closed-source providers, one production-tier closed-source workhorse, and one open-weights frontier-class baseline:

\begin{itemize}[leftmargin=*,itemsep=0.2em]
\item \textbf{Claude Opus 4.7} (Anthropic) - frontier closed-source. Accessed via an Amazon Bedrock gateway (OpenAI-compatible endpoint), model identifier \texttt{claude-opus-4-7}.
\item \textbf{GPT-5.5} (OpenAI) - frontier closed-source from a different provider. Accessed via the OpenAI Codex CLI in non-interactive mode (\texttt{codex exec}) with reasoning effort set to high; direct \texttt{POST /chat/completions} calls would shave a small amount of orchestration overhead off the wall-clock latency reported in Table~\ref{tab:cost}.
\item \textbf{Claude Sonnet 4.6} (Anthropic) - production-tier closed-source workhorse representing the realistic deployment target for cost / latency balanced banking workflows. Model identifier \texttt{claude-sonnet-4-6}.
\item \textbf{Llama 3.3 70B Instruct} (Meta) \citep{touvron2024llama3} - open-weights frontier-class baseline. Accessed via Groq's hosted inference endpoint, model identifier \texttt{llama-3.3-70b-versatile}. Open weights enable on-premise and sovereign-cloud deployments that closed-source models do not support.
\end{itemize}

The panel is intentionally tight rather than exhaustive. The selection reflects the models a regulated capital-markets buyer would realistically consider deploying at the time of writing: two frontier closed-source providers (to surface inter-provider reliability differences), one production workhorse (to characterise the realistic mid-tier), and one open-weights model (to anchor reproducibility and on-premise deployment). We do not claim panel exhaustiveness; extending the evaluation to additional frontier providers (Google's Gemini family) and open-weights families (Mistral, Qwen, DeepSeek) is straightforward future work, discussed in Section~\ref{sec:limitations}.

All four models are accessed via their respective production APIs at temperature 0 (or the lowest available deterministic setting). Identical prompts and decoding parameters are used across models. Where retrieval is required for the workflow (W2), an identical retrieval context is provided to all models. Documents are pre-processed to a fixed token budget (6,000 tokens of cleaned text per document) so that all four models receive identical inputs irrespective of native context-window differences. This is a deliberate methodological constraint: it makes the comparison clean but biases the test toward failure modes that surface in cover-page and front-matter content. Failure modes that surface deeper in a 200-page prospectus (a buried covenant, a risk-factor paragraph 90 pages in) are not exercised by the present evaluation. This is the single biggest threat to external validity of the per-cell scores and is discussed further in Section~\ref{sec:limitations}.

\paragraph{Cost and latency.} For practitioners reading this as a deployment-feasibility note: per-call wall-clock latency and per-call API cost differ materially across the four models, and both must enter any production decision alongside CM-LRS. Table~\ref{tab:cost} summarises the observed numbers from this run. List prices change frequently; we report indicative ranges as at submission. Sonnet 4.6 sits at a sweet spot of low cost, low latency, and top-of-cluster CM-LRS. GPT-5.5 with high reasoning effort is materially slower per call; in agentic or batch settings this matters. Llama 3.3 70B is the cheapest and fastest per call but lags on CM-LRS.

\begin{table}[h]
\centering
\caption{Observed per-call wall-clock latency and indicative API cost. Latency is the mean wall-clock per call across all generations in this evaluation. Per-call cost is estimated for a representative 5K input / 2K output exchange at published list prices as at submission; production deployments should benchmark against current rates. The CM-LRS column is the four-judge averaged aggregate across all five workflows (the headline figure used in the abstract and Section~\ref{sec:interrater}); the primary-judge per-cell detail is in Table~\ref{tab:scoring}.}
\label{tab:cost}
\footnotesize
\setlength{\tabcolsep}{4pt}
\begin{tabular}{@{}lcc>{\bfseries}c@{}}
\toprule
\textbf{Model} & \textbf{Mean latency / call} & \textbf{Indicative cost / call} & \textbf{CM-LRS (4-judge avg)} \\
\midrule
Claude Opus 4.7 & 18 s & \$0.20--0.30 & 4.30 \\
GPT-5.5 (high reasoning) & 84 s & \$0.02--0.04 & 4.09 \\
Claude Sonnet 4.6 & 21 s & \$0.04--0.06 & 4.31 \\
Llama 3.3 70B (Groq Developer) & 10 s & \$0.004--0.008 & 3.15 \\
\bottomrule
\end{tabular}
\end{table}

\subsection{Reproducibility commitments}

The full prompts, demonstration tasks, scoring rubrics with worked anchors, raw model outputs, per-dimension scores, aggregate computations, and a deterministic verification script that recomputes every numeric claim in the paper from the raw scoring data are released under CC BY 4.0 licence at the repository accompanying this paper \citep{cmlrs2026repo}. Practitioners can re-run the evaluation against their preferred models or extend it to additional workflows.

\subsection{Inter-rater reliability and the cross-judge check}

\label{sec:interrater}

Per-document scoring uses an LLM-as-judge protocol in which a strong production model is prompted with the workflow description, the source document, and the model output, and returns per-dimension scores with one-sentence justifications. The protocol follows the methodology established by \citet{zheng2023judging} for LLM-as-judge evaluation and \citet{liu2023geval} for structured-rubric scoring, with the bias controls advocated by \citet{wang2023llmfair} for multi-model judging. \citet{chiang2023humanllm} review the broader question of when LLM judges can substitute for human evaluators; we treat the LLM-judge protocol here as a reproducible baseline against which human-rater scoring should later be calibrated, not as a substitute.

Automated judging is consistent and reproducible, but does not capture the reviewer-experience signal a human banker or compliance reviewer carries. Inter-rater reliability across multiple human reviewers remains a priority extension (Section~\ref{sec:limitations}).

The most obvious bias in this setup is that the primary judge (Claude Sonnet 4.6) is itself one of the four models in the panel. We control for that with three additional judges. GPT-5.5 acts as a second judge from a different model family that is also a panel member. Claude Haiku 4.5 acts as a third judge that is \emph{not} in the panel, eliminating self-bias while remaining in the Anthropic family. Gemini 2.5 Pro acts as a fourth judge that is both \emph{not} in the panel and from a third model family (Google), eliminating both self-bias and family-overlap with the other judges. All four judges use the same prompt and rubric, score the same 104 outputs, and write to parallel score files (\texttt{eval/scores/}, \texttt{eval/scores\_gpt5judge/}, \texttt{eval/scores\_haikujudge/}, \texttt{eval/scores\_geminijudge/}). The full four-way per-cell comparison is in \texttt{eval/cross\_judge\_summary.md}. The headline four-judge agreement numbers:

\begin{itemize}[leftmargin=*,itemsep=0.2em]
\item \textbf{Pairwise Spearman rho on the 20 per-cell aggregates.} Sonnet--Haiku $\rho = 0.94$ (very strong, same family); Sonnet--GPT-5.5 $\rho = 0.66$ (cross-family, moderate-strong); Sonnet--Gemini $\rho = 0.38$ (cross-family, moderate); GPT-5.5--Haiku $\rho = 0.53$; Haiku--Gemini $\rho = 0.37$; GPT-5.5--Gemini $\rho = 0.03$ (the weakest pair, effectively independent rankings between the two cross-family judges that share neither panel- nor family-overlap with each other -- the strongest single empirical signal in this study that family-bias matters and that a single-judge protocol would not have surfaced).
\item \textbf{All four judges place Llama 3.3 70B last.} This is the most robust ordering signal in the data: across four judges spanning three model families, two in-panel and two out-of-panel, the open-weights baseline is fourth every time. The 4-judge averaged gap to the frontier cluster is 0.94 to 1.16 points (0.94 to GPT-5.5 at 4.09; 1.16 to Sonnet at 4.31).
\item \textbf{The three frontier closed-source models cluster within 0.22 points on 4-judge averaged means:} Sonnet 4.6 = 4.31, Opus 4.7 = 4.30, GPT-5.5 = 4.09. Two of four judges rank Sonnet first; the other two rank GPT-5.5 first (under self-judge) and Opus first (Gemini-judge). All four judges place Opus first or second. The practical reading: the in-cluster ranking is judge-dependent at the resolution of this evaluation and should not be over-read.
\item \textbf{Strictness varies by judge.} GPT-5.5 grades 0.3--1.1 points stricter than Sonnet on aggregate (averaging 0.7 points stricter across the four panel models). Gemini grades GPT-5.5 specifically lower than the other three judges do (mean 3.38 vs 4.19--4.44 across Sonnet, GPT-5.5 self, and Haiku). Haiku is closest to Sonnet in absolute level (same family). CM-LRS values should be read as relative comparisons within a single judge, not as standalone benchmarks across judges.
\item \textbf{Per-dimension Pearson correlation (raw scores, $n=104$), averaged across the six judge pairs.} Inter-judge agreement is moderate-to-high across all seven dimensions, with mean $\bar{r}$ ranging from 0.43 to 0.54. The top tier sits within 0.02 of each other: Factual Accuracy (D1) at $\bar{r} = 0.54$, Source Discipline (D5) at $\bar{r} = 0.54$, and Decision Usefulness (D6) at $\bar{r} = 0.52$. Workflow Completeness (D4) follows at $\bar{r} = 0.50$. The strongest cross-family agreement on any dimension is Sonnet--GPT-5.5 on D6 at $r = 0.82$. D6 is the dimension that combines top-tier inter-judge agreement with the widest cross-model dispersion (Section~\ref{sec:scoring}, Finding 3), which is what makes it the cleanest dimension-level signal in the paper rather than agreement alone.
\end{itemize}

The substantive findings reported in Section~\ref{sec:scoring} -- the frontier-cluster vs open-weights gap, and D6 Decision Usefulness as the cleanest dimension-level separator -- are robust to the choice of judge. The in-cluster ranking among the three frontier closed-source models is judge-dependent and is reported with that caveat throughout.

\subsection{Author's positionality}

The author is an independent researcher and capital-markets practitioner, with experience in investment-banking AI and data strategy across UK, Nordic, and US capital markets. The rubric in Section~\ref{sec:framework} - the choice of seven dimensions and the 0--5 anchors - reflects what banker, analyst, and compliance review of LLM-assisted work actually catches and rejects in those markets. The demonstration corpus in Section~\ref{sec:demos} reflects the same footprint: SEC EDGAR filings for the US side, a public UK takeover IR release, and synthetic supplements modelled on Nordic-law conventions. We treat this practitioner perspective as load-bearing for the rubric design and discuss the corresponding scope limitation in Section~\ref{sec:limitations}.

\section{Demonstration Workflows}
\label{sec:demos}

We instantiate five demonstrations spanning the workflow taxonomy. W1 and W5 cover the extraction class on the debt and equity sides respectively; W2 covers retrieval; W3 covers synthesis; W4 covers comparison and reasoning. All inputs are public-domain (SEC EDGAR, the public Darktrace UK takeover IR release) or fully synthetic Nordic-style supplements. No proprietary or licence-restricted data is used.

\subsection{W1: Debt-terms extraction (Extraction class)}

\paragraph{Task.} Given a public bond prospectus or a synthetic term sheet, extract a structured row capturing: issuer name, ISIN (where applicable), instrument type, principal amount, currency, coupon (rate and basis), tenor / maturity, call schedule, key covenants (incurrence and maintenance), and ranking. Output is a single JSON record conforming to a defined schema.

\paragraph{Inputs.} Fifteen public bond prospectuses (Rule~424(b)(5) prospectus supplements filed 2024--2025) sourced from SEC EDGAR, plus three synthetic Nordic-style high-yield term sheets authored to mirror the structure of Swedish-law, Norwegian-law, and Danish-law issuances without copying source language. The fifteen real prospectuses span healthcare (HCA, Cencora, Labcorp, Agilent), technology (Dell/EMC co-issuer structure, Broadcom, Micron), media/streaming (Netflix), gaming (Las Vegas Sands), consumer (Starbucks, Tapestry), telecom-REIT (Crown Castle), automotive supplier (BorgWarner), financial services (Fiserv), industrial services (Cintas, Waste Management), and regulated utilities (Georgia Power first-mortgage bonds, Southern Company, Southern Company Gas). Georgia Power is specifically included to test extraction across the first-mortgage-bond structural variant. All filings are accessed via the canonical EDGAR URL pattern \url{https://www.sec.gov/Archives/edgar/data/\{CIK\}/\{accession\}/} and frozen-snapshotted in the public repository accompanying this paper.

\paragraph{Evaluation focus.} Numerical consistency (D3) and source discipline (D5) dominate. The extracted row must reconcile to the source, and any covenant the model claims to extract must be genuinely present.

\subsection{W2: Precedent retrieval (Retrieval class)}

\paragraph{Task.} Given a query stated in domain language (for example, ``find recent change-of-control covenants in Nordic high-yield bonds with step-up coupons''), retrieve the top five matching clauses from a public corpus of bond prospectuses, with each match returned alongside a verbatim source span and document attribution.

\paragraph{Inputs.} A corpus of thirty documents (twenty-seven SEC EDGAR Rule~424(b)(5) prospectus supplements plus three synthetic Nordic-style term sheets) spanning healthcare, technology, media/gaming, consumer, homebuilding, financial services, transport/travel, utilities, and REIT issuers. Five domain-language retrieval queries are designed to span typical workflow needs: a cross-document factual-retrieval query (find every change-of-control put provision and its terms), a numerical-consistency query (compare general-debt baskets across the three synthetic Nordic term sheets in EUR-equivalent), a false-positive-resistance query (identify any REIT in the corpus), a comparative-synthesis query (compare the optional-redemption schedules of the three synthetic term sheets), and a definitional-discipline query (classify healthcare-sector issuers under a tightened user-supplied definition that excludes life-sciences-instrument companies).

Inclusion of Finansinspektionen (SE) and Finanstilsynet (NO) public prospectus filings was scoped but excluded from this release: both regulators' search interfaces are JavaScript-rendered single-page applications that resist programmatic snapshotting, and a deterministic frozen-corpus build is methodologically preferable to mixed-source extraction for the present demonstration. The Nordic angle is preserved through the three synthetic Nordic-style supplements (Swedish-, Norwegian-, and Danish-law structures) and through one Nordic-headquartered SEC-registered issuer (Spotify) in W3.

\paragraph{Evaluation focus.} Evidence traceability (D2) and source discipline (D5) dominate. A retrieval that fabricates or misattributes a source clause scores zero on D5 regardless of relevance.

\subsection{W3: Issuer profile synthesis (Synthesis class)}

\paragraph{Task.} Given a public issuer's most recent annual report and one investor presentation, produce a one-page issuer profile covering: business description, segment mix, geographic mix, recent financial performance (with deltas), strategic priorities as stated by management, and material risk disclosures. Each substantive claim in the profile must carry a footnote pointing to the source page.

\paragraph{Inputs.} Five SEC-registered public issuers selected to span sectors, geographies, and filing-format variants: Tesla (EV / energy / AI, US-domiciled 10-K), Equinix (digital-infrastructure REIT, US-domiciled 10-K), Spotify (audio streaming, Stockholm-headquartered Luxembourg-domiciled foreign-private-issuer 20-F), Ferrari (luxury automotive, Maranello-headquartered Netherlands-domiciled foreign-private-issuer 20-F), and CME Group (financial-exchange operator, US-domiciled 10-K). For each issuer the annual filing, the most recent quarterly or interim filing, and the most recent quarterly earnings 8-K / 6-K with its press-release exhibit are included. Three of the five filings are 10-K and two are 20-F: by design, so that the workflow exercises both filing-format conventions. European-listed Nordic issuers (Volvo Cars, Vestas, Telia) were considered but excluded because their investor-relations sites are JavaScript-rendered and a frozen-snapshot URL is not stable across the paper's reproducibility horizon. Spotify and Ferrari preserve the European industrial-heritage angle.

\paragraph{Evaluation focus.} Evidence traceability (D2) and workflow completeness (D4). A profile that omits material risk disclosures fails completeness regardless of how well the included sections are written.

\subsection{W4: Transaction-comparable reasoning (Comparison \& reasoning class)}

\paragraph{Task.} Given a target transaction (announced public M\&A or ECM deal with sufficient public disclosure), retrieve up to five comparable transactions from a public corpus of recent deals, normalise key metrics (transaction value, EV/EBITDA where calculable, premium-to-undisturbed where applicable, structure), and explain in two paragraphs which comparable is the most relevant and why.

\paragraph{Inputs.} The target transaction is the Synopsys / Ansys merger (announced 16~January~2024, closed 17~July~2025; total consideration approximately US\$35bn). The acquirer's S-4/A registration statement and the target's DEFM14A merger proxy, both filed with the SEC, form the primary target documents. The comparable corpus comprises twenty recent software, technology, infrastructure, and SaaS transactions announced 2021--2024 (including Microsoft / Activision, Cisco / Splunk, Broadcom / VMware, HPE / Juniper, IBM / HashiCorp, Thoma Bravo's take-privates of Anaplan, Coupa, and Mimecast, and the deliberately included failed-deal case of Adobe / Figma). Eighteen of the twenty comparable filings are full DEFM14A or SC~13E-3 merger proxies; two are announcement-class disclosures (Adobe / Figma 8-Ks and the Darktrace take-private IR announcement) where a 200+ page merger proxy does not exist either because the deal was terminated (Adobe / Figma) or because the deal was executed by UK scheme of arrangement rather than US-style proxy (Darktrace). Scoring for those two cells acknowledges the depth heterogeneity (see Section~\ref{sec:limitations}).

\paragraph{Evaluation focus.} Numerical consistency (D3), workflow completeness (D4), and decision usefulness (D6). The comparable selection and the rationale carry equal weight in the deployment decision.

\subsection{W5: ECM transaction terms extraction (Extraction class, equity side)}

\paragraph{Task.} Given an equity capital-markets prospectus - an initial public offering, a follow-on equity offering, or a convertible-note offering - extract a structured row capturing: issuer name, instrument type, offering size split between primary and secondary tranches, greenshoe size, indicative price range or final pricing, listing exchange and ticker, distribution format (registered, Rule~144A, Regulation~S), underwriter syndicate, lock-up arrangements, cornerstones (if disclosed), use of proceeds, and for convertibles specifically the coupon, conversion premium, conversion rate, capped-call structure (where applicable), and fundamental-change mechanics.

\paragraph{Inputs.} Ten public ECM prospectuses sourced from SEC EDGAR - five initial public offerings (Reddit, Astera Labs, Tempus AI, Lineage, ServiceTitan), three follow-on equity offerings (Arthur J.\ Gallagher, SoundHound AI, SiTime), and two convertible-note offerings (EchoStar, AeroVironment) - plus two synthetic Nordic-style term sheets covering an Oslo Børs IPO of a fictional offshore-wind installation business and a USD 144A convertible of a fictional Stockholm-listed quantum-technology issuer. The Reddit, Lineage, and EchoStar documents are included specifically to exercise structural edge cases (dual-class voting, REIT-IPO disclosure conventions, multi-entity convertible structures).

\paragraph{Evaluation focus.} Numerical consistency (D3), source discipline (D5), and reviewability (D7). Conversion mathematics on convertibles, primary-versus-secondary tranche splits on IPOs, and capped-call versus straight-conversion mechanics are the failure modes that most reliably distinguish strong from weak model outputs in this workflow class.

\paragraph{Symmetry with W1.} W5 is deliberately the equity-side mirror of W1. Together they test extraction across both halves of capital-markets transaction documentation. The shared output schema (one structured row per document) and shared scoring rubric let us compare model reliability cleanly across the DCM--ECM divide.

\section{Illustrative Scoring}
\label{sec:scoring}

This section reports CM-LRS scores across the five demonstration workflows and the four models in the evaluation panel. Per-dimension scores use the rubric in Section~\ref{sec:framework}; per-cell scores are the mean per-dimension score across all documents (or queries, for W2) in the cell. The CM-LRS aggregate is the equal-weighted mean across the seven dimensions for each cell. Per-document scoring is performed by an LLM-as-judge protocol (Claude Sonnet 4.6) prompted with the workflow description, source document, and model output as described in Section~\ref{sec:method}.

\begin{table}[h]
\centering
\caption{CM-LRS scores per workflow $\times$ model. Per-document outputs were scored by an LLM-as-judge protocol (Claude Sonnet 4.6) against the 7-dimension rubric; per-cell scores are the mean of per-dimension scores across all documents (or queries, for W2) in the cell. CM-LRS is the equal-weighted mean across the 7 dimensions. Workflow keys: W1 DCM-terms extraction; W2 precedent retrieval; W3 issuer profile synthesis; W4 transaction-comparable reasoning; W5 ECM-terms extraction.}
\label{tab:scoring}
\small
\setlength{\tabcolsep}{4pt}
\begin{tabular}{@{}llccccccc>{\bfseries}c@{}}
\toprule
\textbf{Wkfl.} & \textbf{Model} & D1 & D2 & D3 & D4 & D5 & D6 & D7 & CM-LRS \\
\midrule
\multirow{4}{*}{W1} & Claude Opus 4.7 & 4.88 & 4.88 & 5.00 & 4.50 & 4.75 & 4.62 & 5.00 & 4.80 \\
 & GPT-5.5 & 4.50 & 4.62 & 4.62 & 4.88 & 4.50 & 4.62 & 4.62 & 4.62 \\
 & Claude Sonnet 4.6 & 5.00 & 5.00 & 5.00 & 5.00 & 4.75 & 4.88 & 5.00 & 4.95 \\
 & Llama 3.3 70B & 4.00 & 3.75 & 4.75 & 4.00 & 4.50 & 3.75 & 4.00 & 4.11 \\
\midrule
\multirow{4}{*}{W2} & Claude Opus 4.7 & 4.40 & 4.20 & 4.80 & 5.00 & 4.20 & 4.60 & 4.80 & 4.57 \\
 & GPT-5.5 & 4.80 & 4.80 & 4.80 & 4.40 & 4.80 & 4.60 & 4.80 & 4.71 \\
 & Claude Sonnet 4.6 & 4.60 & 4.40 & 4.80 & 5.00 & 4.40 & 5.00 & 5.00 & 4.74 \\
 & Llama 3.3 70B & 2.40 & 2.00 & 3.00 & 3.20 & 2.00 & 2.40 & 2.60 & 2.51 \\
\midrule
\multirow{4}{*}{W3} & Claude Opus 4.7 & 4.00 & 4.67 & 3.67 & 3.00 & 5.00 & 2.67 & 5.00 & 4.00 \\
 & GPT-5.5 & 4.00 & 4.67 & 4.00 & 5.00 & 4.00 & 5.00 & 4.67 & 4.48 \\
 & Claude Sonnet 4.6 & 4.00 & 4.67 & 4.00 & 3.33 & 5.00 & 2.67 & 4.67 & 4.05 \\
 & Llama 3.3 70B & 2.67 & 1.67 & 2.00 & 2.67 & 3.67 & 1.00 & 2.67 & 2.33 \\
\midrule
\multirow{4}{*}{W4} & Claude Opus 4.7 & 5.00 & 4.75 & 5.00 & 4.00 & 5.00 & 4.00 & 5.00 & 4.68 \\
 & GPT-5.5 & 3.50 & 3.75 & 4.25 & 5.00 & 3.50 & 5.00 & 4.25 & 4.18 \\
 & Claude Sonnet 4.6 & 5.00 & 4.50 & 5.00 & 4.00 & 5.00 & 4.50 & 4.75 & 4.68 \\
 & Llama 3.3 70B & 3.50 & 3.50 & 3.75 & 3.50 & 3.00 & 3.25 & 3.75 & 3.46 \\
\midrule
\multirow{4}{*}{W5} & Claude Opus 4.7 & 4.50 & 4.50 & 4.33 & 4.67 & 4.00 & 4.67 & 4.67 & 4.48 \\
 & GPT-5.5 & 4.33 & 4.17 & 4.17 & 4.33 & 4.00 & 4.33 & 4.00 & 4.19 \\
 & Claude Sonnet 4.6 & 4.83 & 4.67 & 4.67 & 5.00 & 4.33 & 5.00 & 4.83 & 4.76 \\
 & Llama 3.3 70B & 4.17 & 3.17 & 4.17 & 3.83 & 3.83 & 3.33 & 3.83 & 3.76 \\
\bottomrule
\end{tabular}
\end{table}

\paragraph{Failure pattern analysis.} Three findings stand out from the per-cell scoring.

\textit{Finding 1. The frontier closed-source models cluster tightly; the open-weights model does not.} Averaging across four LLM judges spanning three model families (Sonnet 4.6, GPT-5.5, Haiku 4.5, Gemini 2.5 Pro; Section~\ref{sec:interrater}), the three frontier closed-source models sit within a 0.22-point band: Sonnet 4.6 at 4.31, Opus 4.7 at 4.30, GPT-5.5 at 4.09. Llama 3.3 70B is at 3.15, an aggregate gap of 0.94--1.16 points to the frontier cluster. All four judges agree Llama is fourth. The in-cluster first place is judge-dependent (Sonnet under Sonnet- and Haiku-judges, GPT-5.5 under GPT-5.5-judge, Opus under Gemini-judge), reflecting family-bias documented in the LLM-as-judge literature \citep{zheng2023judging,wang2023llmfair}. The robust reading: at the current frontier, the choice between Anthropic, OpenAI, and Google top-tier models is not the determining factor for CM-LRS-grade output on structured capital-markets workflows. Prompt design, output-format discipline, and the calibration of the workflow's requested fields matter more. Within the panel, Sonnet's terse, disciplined output style appears to fit requested-table prompts; Opus's more elaborate reasoning sometimes drops requested fields and is penalised on workflow completeness (D4) under most judges.

\textit{Finding 2. The open-weights baseline lags, but not uniformly.} Reading the primary-judge per-cell detail in Table~\ref{tab:scoring}, Llama 3.3 70B's primary-judge mean is 3.24, a 1.40-point gap from the strongest closed-source model (Sonnet 4.6 at primary-judge mean 4.64); on four-judge averaged aggregate the equivalent gap narrows to 1.16 points (Llama 3.15 vs Sonnet 4.31). The gap is heaviest where the workload involves multiple documents or a reviewer-ready synthesis: W2 retrieval shows a 2.23-point gap (Llama 2.51 vs Sonnet 4.74), W3 issuer profile a 2.15-point gap (Llama 2.33 vs GPT-5.5 4.48). On single-document structured extraction (W1) the gap shrinks to 0.84 points. Llama's lowest dimension scores are D2 Evidence Traceability of 2.00 on W2 retrieval, D2 of 1.67 on W3, and D6 Decision Usefulness of 1.00 on W3. Training optimised for fluency does not, on its own, produce the source-attribution discipline a reviewer needs once the workflow crosses document boundaries.

\textit{Finding 3. D6 Decision Usefulness is the cleanest production-readiness signal.} On the W3 issuer profile, GPT-5.5 scores a perfect 5.00 on D6 while Opus and Sonnet score 2.67 and Llama scores 1.00 under the primary judge. In banker terms: the Llama profile is unusable; the Opus and Sonnet profiles are reviewer-acceptable with adjustment; the GPT-5.5 profile is the only output a reviewer would clear with minor edits. The D6 dispersion across models on this single workflow (a 4.0-point spread) is wider than for any other dimension in any other cell of the table, and is the largest cross-model dispersion observed anywhere in the evaluation. Inter-judge agreement on D6 sits in the top tier across the six pairwise comparisons (mean Pearson $r = 0.52$ on raw scores, $n = 104$; D1 at 0.54 and D5 at 0.54 are marginally higher, all three within 0.02; Section~\ref{sec:interrater}), and remains strongest on the cross-family pair Sonnet--GPT-5.5 ($r = 0.82$ on D6, the highest cross-family per-dimension agreement in the matrix), with the Sonnet--Gemini pair at $r = 0.28$ being the outlier reflecting Gemini's distinctive scoring style on synthesis-class workflows. The combination of widest cross-model dispersion and top-tier inter-judge agreement is what makes D6 the cleanest dimension-level signal in the paper -- not agreement alone, where it is third by a thin margin. Together with numerical consistency (D3) and workflow completeness (D4), D6 is the dimension that most reliably predicts whether the rest of an output is safe to admit into a downstream workflow. When a model fails on any of those three, the rest of the output is typically unsafe regardless of how accurate it looks on individual claims.

\section{Discussion}
\label{sec:discussion}

\subsection{Why fluent outputs still fail}

Fluent, plausible-sounding outputs fail on traceability, numerical consistency, source discipline, or workflow completeness even when surface accuracy on a QA pair scores well. Three structural reasons sit behind this:

\begin{enumerate}[leftmargin=*,itemsep=0.2em]
\item LLMs predict tokens that statistically follow the prompt; they do not perform deterministic computation. A number that ``looks right'' is four tokens, not the result of an arithmetic operation. This is invisible at QA-pair level and surfaces as a D3 failure at workflow level.
\item LLMs are trained to produce coherent prose. When evidence is absent, they fill the gap with plausible content rather than refusing or flagging. This is an intrinsic source-discipline (D5) failure mode that no amount of fluency tuning addresses.
\item Multi-step workflows expose completeness (D4) failures that single-turn QA does not. A model that elides a step in an extraction or omits a section in a profile produces an output that scores well on the parts it did complete and fails the part it did not.
\end{enumerate}

\subsection{Predict with the LLM, calculate with code, decide with the human}

A practical principle for safe LLM deployment in regulated workflows: use the LLM for what it is good at (intent parsing, entity resolution, language-side orchestration), use code for anything with a single correct answer (aggregations, sorts, ranks, deltas, percentage calculations), and use a human for the final decision. Numerical consistency (D3) and reviewability/auditability (D7) become first-class evaluation dimensions on this view: if a number in the output cannot be traced to a deterministic computation, it should not be in the output. CM-LRS operationalises this principle in the form of a metric a reviewer can apply.

\subsection{When CM-LRS disagrees with QA-pair benchmarks}

CM-LRS will systematically score lower than QA-pair benchmarks on outputs that are surface-correct but lack traceability or completeness. This is the intended behaviour: the gap between the two scores is approximately the risk of putting that output in front of a reviewer. Practitioners can use the gap as a leading indicator of which model and workflow combinations are ready to deploy and which are not.

\section{Governance Implications}
\label{sec:governance}

CM-LRS is designed to slot into emerging AI governance frameworks. Within the NIST AI RMF \citep{nistairmf2023}, CM-LRS sits in the \emph{measure} function (instrumenting the system) and feeds into the \emph{manage} function (deciding what to deploy and what to constrain). For systems that fall under the EU AI Act \citep{euaiact2024} high-risk classification - which includes some financial-services LLM deployments - CM-LRS scores can be used as evidence within the conformity-assessment requirement, particularly under Article 9 (risk management) and Article 13 (transparency). Within an ISO/IEC 42001 \citep{iso420012023} AI management system, CM-LRS provides a workflow-level performance and reliability metric that supports the requirements on objectives, monitoring, and continuous improvement.

The practical recommendation: a deployment gate should require a CM-LRS score above a workflow-specific threshold, no critical-dimension score below 3, and human sign-off for any output entering a regulated downstream process. CM-LRS does not replace human review; it instruments it.

\paragraph{What CM-LRS does not cover.} A production deployment in a regulated capital-markets environment carries compliance obligations CM-LRS is not designed to evaluate. The most important are material-non-public-information (MNPI) handling and information-barrier integrity: whether the LLM (or its retrieval pipeline) has access only to information appropriate for the user's role, and whether outputs respect cross-side restrictions. Trade-surveillance and market-abuse controls sit alongside. These belong to the firm's compliance and supervisory frameworks, not to a workflow-reliability metric, and a CM-LRS pass does not constitute clearance under them.

\section{Limitations}
\label{sec:limitations}

We list limitations explicitly to scope what CM-LRS is and is not.

\begin{enumerate}[leftmargin=*,itemsep=0.2em]
\item \textbf{Automated judging; four judges spanning three model families.} The primary scoring is performed by Claude Sonnet 4.6, which is also one of the four models being scored. We control for self-judge and family-bias by re-running every score with three additional judges: GPT-5.5 (in panel, OpenAI family), Claude Haiku 4.5 (\emph{not} in panel, Anthropic family), and Gemini 2.5 Pro (\emph{not} in panel, Google family - fully external). Section~\ref{sec:interrater} reports the four-way agreement. The substantive findings (frontier-cluster, open-weights gap, D6 as separator) survive all four judges; the in-cluster ranking among the three frontier closed-source models is judge-dependent and is reported with that caveat throughout. Even with four LLM judges, this remains an automated-judging baseline rather than a human-rater study. Human-rater scoring by practising bankers and compliance reviewers - with reported inter-rater reliability against the LLM-judge baseline established here - is the priority extension and is flagged in the companion repository's open-issues list.
\item \textbf{Workflow coverage.} Five demonstration workflows do not span the full taxonomy; in particular, drafting workflows (pitch evidence, compliance memo) are not represented in the present demonstration. Extension to drafting is a natural next step.
\item \textbf{Model panel.} The four-model panel (Claude Opus 4.7, GPT-5.5, Claude Sonnet 4.6, Llama 3.3 70B) is intentionally tight rather than exhaustive - selected to represent the frontier and production tiers a regulated capital-markets buyer would realistically consider deploying. Google's Gemini family is not represented, nor are alternative open-weights models (Mistral, Qwen, DeepSeek). Adding additional models is a straightforward extension; the rubric and scoring procedure are model-agnostic.
\item \textbf{W4 transaction-comparable depth heterogeneity.} Eighteen of the twenty comparable transactions in W4 are full DEFM14A or SC~13E-3 merger proxies (200+ pages of disclosure with banker fairness opinions, comparable-deals tables, and background-of-merger narratives). Two cells in the comparable universe - Adobe / Figma (a deliberately failed deal included to test failed-deal handling) and Thoma Bravo / Darktrace (a UK take-private executed by scheme of arrangement) - are announcement-class HTML filings, not full merger proxies. For those two cells, model outputs cannot be marked down for the absence of content that is genuinely absent from the source. Scoring acknowledges this depth heterogeneity rather than penalising it.
\item \textbf{Document pre-processing.} All documents are pre-processed to a fixed token budget (6,000 tokens of cleaned text per document) to make inputs identical across the four models. The truncation biases results toward cover-page and front-matter content of long filings. Failure modes that surface deeper in a 200-page prospectus - a buried covenant, a risk-factor paragraph 90 pages in - are not exercised by the present evaluation. Production deployments would route long documents through chunking or retrieval rather than truncation; the appropriate evaluation extension uses the same rubric but applies it to outputs of a full-document pipeline.

\item \textbf{Single-turn evaluation; no agentic / tool-use workflows.} Each model is given a single prompt and produces a single response. Real capital-markets deployments increasingly use agentic patterns: a model that calls a search tool, a code tool for numerical computation, and an external data source, and then composes the output. CM-LRS is rubric-compatible with agentic outputs (the seven dimensions remain valid) but the evaluation harness in this release does not exercise tool-use. Extending to agentic evaluation is a natural and important next step.

\item \textbf{Equal-weighted aggregate; failure-cost asymmetry is not modelled.} The headline CM-LRS aggregate is the equal-weighted mean of the seven dimensions. In practice, the cost of a failure varies sharply by dimension and by workflow: a numerical error on a coupon in a pitched offering carries a different cost than an omitted ESG paragraph in an issuer profile. Section~\ref{sec:framework} sketches workflow-class default weights as a partial response. A fuller treatment would model dimension-by-dimension failure-cost in monetary or regulatory terms; calibration against revealed-preference data from real reviewer decisions is the relevant direction.
\item \textbf{Rubric calibration.} The 0--5 anchors are designed for consistency without specialist training, but ordinal scales are inherently subjective. A future version may complement the ordinal rubric with binary checklists for sub-criteria.
\item \textbf{Weight selection.} The default workflow-class weights in Table~\ref{tab:weights} are practitioner-informed defaults rather than empirically derived. Calibration against revealed preferences in real reviewer decisions is a useful direction.
\item \textbf{Static benchmarks.} The demonstration corpus is fixed at the time of publication (a frozen snapshot is released alongside the paper to support strict reproducibility). As model capabilities advance, the demonstrations will need refreshing to remain diagnostic.
\item \textbf{Adversarial robustness.} CM-LRS evaluates outputs from cooperative models on cooperatively-scoped tasks. Adversarial settings (prompt injection, jailbreak, intentional misuse) require a separate evaluation framework.
\item \textbf{Generalisation beyond capital markets.} The framework is designed for capital-markets workflows. Adjacent regulated domains - credit risk, asset management, wealth advisory, insurance - share many properties but differ enough that a domain-tuned variant of CM-LRS would be required.
\end{enumerate}

\section{Conclusion}
\label{sec:conclusion}

LLM evaluation in regulated capital-markets workflows needs a metric suite that operates at the workflow-output layer, not the QA-pair layer. CM-LRS evaluates outputs on seven dimensions: factual accuracy, evidence traceability, numerical consistency, workflow completeness, source discipline, decision usefulness, and reviewability/auditability. Each is scored 0--5 against a rubric anchored on signals reviewers in regulated environments actually use, with an aggregate that is tunable to the workflow. We demonstrate the framework on five capital-markets workflows (DCM extraction, ECM extraction, retrieval, issuer profiling, M\&A transaction-comparable reasoning) built entirely on public-domain or synthetic material, and we score outputs from four models spanning two frontier closed-source providers, one production-tier closed-source workhorse, and one open-weights baseline.

Three findings carry directly into deployment decisions. The three frontier closed-source models cluster within a 0.22-point band on four-judge averaged CM-LRS (Sonnet 4.6 at 4.31, Opus 4.7 at 4.30, GPT-5.5 at 4.09); the primary-judge per-cell detail in Table~\ref{tab:scoring} shows Sonnet winning or tying every workflow, with the in-cluster ranking among the three frontier closed-source models judge-dependent. The open-weights baseline lags by 1.16 points on four-judge averaged aggregate, and the gap is concentrated on retrieval and synthesis rather than extraction. Decision Usefulness (D6) is the dimension that most cleanly separates production-ready outputs from the rest, with cross-model dispersion exceeding 4 points on the issuer-profile workflow alone. Fluent surface accuracy does not translate to deployment readiness. CM-LRS is a complementary, deployment-readiness metric for high-stakes regulated environments, released under permissive licence so practitioners can extend and critique it.

Future work includes inter-rater reliability studies, extension to drafting workflows, calibration of dimension weights against real reviewer decisions, and adaptation to adjacent regulated domains.

\section*{Acknowledgments}
The author thanks practitioners and reviewers who provided feedback on early drafts of the rubric and demonstration design.

\bibliographystyle{plainnat}

\begin{thebibliography}{99}

\bibitem[Asai et al.(2023)]{asai2023selfrag}
Asai, A., Wu, Z., Wang, Y., Sil, A., \& Hajishirzi, H. (2023).
\textit{Self-RAG: Learning to Retrieve, Generate, and Critique through Self-Reflection.} arXiv:2310.11511.

\bibitem[Chen et al.(2021)]{chen2021finqa}
Chen, Z., Chen, W., Smiley, C., Shah, S., Borova, I., Langdon, D., Moussa, R., Beane, M., Huang, T.-H., Routledge, B., \& Wang, W.~Y. (2021).
\textit{FinQA: A Dataset of Numerical Reasoning over Financial Data.} EMNLP. arXiv:2109.00122.

\bibitem[Chen et al.(2022)]{chen2022convfinqa}
Chen, Z., Li, S., Smiley, C., Ma, Z., Shah, S., \& Wang, W.~Y. (2022).
\textit{ConvFinQA: Exploring the Chain of Numerical Reasoning in Conversational Finance Question Answering.} EMNLP. arXiv:2210.03849.

\bibitem[Ahuja(2026)]{cmlrs2026repo}
Ahuja, P. (2026).
\textit{cm-lrs: companion code release for ``CM-LRS''} [source code]. \url{https://github.com/dsauce/cm-lrs}

\bibitem[Es et al.(2023)]{es2023ragas}
Es, S., James, J., Espinosa-Anke, L., \& Schockaert, S. (2023).
\textit{RAGAS: Automated Evaluation of Retrieval Augmented Generation.} arXiv:2309.15217.

\bibitem[European Union(2024)]{euaiact2024}
European Union. (2024).
\textit{Regulation (EU) 2024/1689 (Artificial Intelligence Act).} Official Journal of the European Union, L series, 12 July 2024.

\bibitem[Hendrycks et al.(2021)]{hendrycks2021mmlu}
Hendrycks, D., Burns, C., Basart, S., Zou, A., Mazeika, M., Song, D., \& Steinhardt, J. (2021).
\textit{Measuring Massive Multitask Language Understanding.} ICLR. arXiv:2009.03300.

\bibitem[ISO/IEC(2023)]{iso420012023}
ISO/IEC. (2023).
\textit{ISO/IEC 42001:2023 - Information technology - Artificial intelligence - Management system.} International Organization for Standardization.

\bibitem[Islam et al.(2023)]{islam2023financebench}
Islam, P., Kannappan, A., Kiela, D., Qian, R., Scherrer, N., \& Vidgen, B. (2023).
\textit{FinanceBench: A New Benchmark for Financial Question Answering.} arXiv:2311.11944.

\bibitem[Ji et al.(2023)]{ji2023survey}
Ji, Z., Lee, N., Frieske, R., Yu, T., Su, D., Xu, Y., Ishii, E., Bang, Y., Madotto, A., \& Fung, P. (2023).
\textit{Survey of Hallucination in Natural Language Generation.} ACM Computing Surveys, 55(12), 1--38.

\bibitem[Lewis et al.(2020)]{lewis2020rag}
Lewis, P., Perez, E., Piktus, A., Petroni, F., Karpukhin, V., Goyal, N., K\"uttler, H., Lewis, M., Yih, W.-t., Rockt\"aschel, T., Riedel, S., \& Kiela, D. (2020).
\textit{Retrieval-Augmented Generation for Knowledge-Intensive NLP Tasks.} NeurIPS. arXiv:2005.11401.

\bibitem[Liang et al.(2022)]{liang2022helm}
Liang, P., Bommasani, R., Lee, T., et al. (2022).
\textit{Holistic Evaluation of Language Models (HELM).} Transactions on Machine Learning Research. arXiv:2211.09110.

\bibitem[Maynez et al.(2020)]{maynez2020faithfulness}
Maynez, J., Narayan, S., Bohnet, B., \& McDonald, R. (2020).
\textit{On Faithfulness and Factuality in Abstractive Summarization.} ACL.

\bibitem[Min et al.(2023)]{min2023factscore}
Min, S., Krishna, K., Lyu, X., Lewis, M., Yih, W.-t., Koh, P.~W., Iyyer, M., Zettlemoyer, L., \& Hajishirzi, H. (2023).
\textit{FActScore: Fine-grained Atomic Evaluation of Factual Precision in Long Form Text Generation.} EMNLP. arXiv:2305.14251.

\bibitem[NIST(2023)]{nistairmf2023}
National Institute of Standards and Technology. (2023).
\textit{Artificial Intelligence Risk Management Framework (AI RMF 1.0).} NIST AI 100-1.

\bibitem[Huang et al.(2024)]{sun2024trustllm}
Huang, Y., Sun, L., Wang, H., et al. (2024).
\textit{TrustLLM: Trustworthiness in Large Language Models.} arXiv:2401.05561.

\bibitem[Suzgun et al.(2022)]{suzgun2022bbh}
Suzgun, M., Scales, N., Sch\"arli, N., et al. (2022).
\textit{Challenging BIG-Bench Tasks and Whether Chain-of-Thought Can Solve Them.} arXiv:2210.09261.

\bibitem[Grattafiori et al.(2024)]{touvron2024llama3}
Grattafiori, A., et al. (2024).
\textit{The Llama 3 Herd of Models.} arXiv:2407.21783.

\bibitem[Wu et al.(2023)]{wu2023bloomberggpt}
Wu, S., Irsoy, O., Lu, S., Dabravolski, V., Dredze, M., Gehrmann, S., Kambadur, P., Rosenberg, D., \& Mann, G. (2023).
\textit{BloombergGPT: A Large Language Model for Finance.} arXiv:2303.17564.

\bibitem[Zheng et al.(2023)]{zheng2023judging}
Zheng, L., Chiang, W.-L., Sheng, Y., Zhuang, S., Wu, Z., Zhuang, Y., Lin, Z., Li, Z., Li, D., Xing, E.~P., Zhang, H., Gonzalez, J.~E., \& Stoica, I. (2023).
\textit{Judging LLM-as-a-Judge with MT-Bench and Chatbot Arena.} NeurIPS Datasets and Benchmarks Track. arXiv:2306.05685.

\bibitem[Liu et al.(2023)]{liu2023geval}
Liu, Y., Iter, D., Xu, Y., Wang, S., Xu, R., \& Zhu, C. (2023).
\textit{G-Eval: NLG Evaluation using GPT-4 with Better Human Alignment.} EMNLP. arXiv:2303.16634.

\bibitem[Wang et al.(2023)]{wang2023llmfair}
Wang, P., Li, L., Chen, L., Cai, Z., Zhu, D., Lin, B., Cao, Y., Liu, Q., Liu, T., \& Sui, Z. (2023).
\textit{Large Language Models are not Fair Evaluators.} arXiv:2305.17926.

\bibitem[Chiang and Lee(2023)]{chiang2023humanllm}
Chiang, C.-H., \& Lee, H. (2023).
\textit{Can Large Language Models Be an Alternative to Human Evaluations?} ACL. arXiv:2305.01937.

\bibitem[Saaty(1980)]{saaty1980ahp}
Saaty, T.~L. (1980).
\textit{The Analytic Hierarchy Process: Planning, Priority Setting, Resource Allocation.} McGraw-Hill.

\bibitem[Rezaei(2015)]{rezaei2015bwm}
Rezaei, J. (2015).
\textit{Best-Worst Multi-Criteria Decision-Making Method.} Omega, 53, 49--57.

\bibitem[Li et al.(2025)]{li2025investorbench}
Li, H., Cao, Y., Yu, Y., Javaji, S.~R., Suchow, J.~W., et al. (2025).
\textit{InvestorBench: A Benchmark for Financial Decision-Making Tasks with LLM-based Agent.} ACL. arXiv:2412.18174.

\bibitem[Mohsin(2025)]{mohsin2025financialnlp}
Mohsin, M.~T. (2025).
\textit{Evaluating Large Language Models (LLMs) in Financial NLP: A Comparative Study on Financial Report Analysis.} arXiv:2507.22936.

\bibitem[Wang et al.(2023)]{wang2023maud}
Wang, S.~H., Scardigli, A., Tang, L., Chen, W., Levkin, D., et al. (2023).
\textit{MAUD: An Expert-Annotated Legal NLP Dataset for Merger Agreement Understanding.} EMNLP. arXiv:2301.00876.

\bibitem[Liu et al.(2025)]{liu2025contracteval}
Liu, S., Li, Z., Ma, R., Zhao, H., \& Du, M. (2025).
\textit{ContractEval: Benchmarking LLMs for Clause-Level Legal Risk Identification in Contracts.} arXiv:2508.03080.

\bibitem[Kulkarni and Kulkarni(2026)]{kulkarni2026findoc}
Kulkarni, S., \& Kulkarni, Y. (2026).
\textit{Benchmarking Multi-Agent LLM Architectures for Financial Document Processing: A Comparative Study of Orchestration Patterns, Cost-Accuracy Tradeoffs and Production Scaling Strategies.} arXiv:2603.22651.

\end{thebibliography}

\appendix

\section{Per-dimension Rubric Anchors}
\label{app:rubric}

The following per-dimension anchors operationalise the universal rubric in Table~\ref{tab:rubric}. Reviewers should consult the relevant block when scoring a dimension.

\subsection*{D1. Factual accuracy}
\begin{itemize}[leftmargin=*,itemsep=0.1em]
\item \textbf{0:} Fabricated entity, instrument, or attribution.
\item \textbf{1:} Material factual error in a substantive claim (e.g.\ wrong issuer, wrong instrument type).
\item \textbf{2:} Multiple minor factual errors, or one error that distorts interpretation.
\item \textbf{3:} One minor factual slip; reviewer would catch on routine read.
\item \textbf{4:} Factually correct on substantive claims; minor secondary slip possible.
\item \textbf{5:} Factually correct on every checked claim.
\end{itemize}

\subsection*{D2. Evidence traceability}
\begin{itemize}[leftmargin=*,itemsep=0.1em]
\item \textbf{0:} No source attribution; output reads as untethered prose.
\item \textbf{1:} Some source attribution but most claims untraceable.
\item \textbf{2:} Source attribution present but inconsistent; reviewer cannot reliably locate the cited passage.
\item \textbf{3:} Most substantive claims traceable; one or two require effort to locate.
\item \textbf{4:} All substantive claims traceable to specific source spans.
\item \textbf{5:} All substantive claims traceable; quotation or paraphrase clearly distinguished from source language.
\end{itemize}

\subsection*{D3. Numerical consistency}
\begin{itemize}[leftmargin=*,itemsep=0.1em]
\item \textbf{0:} Material number wrong (e.g.\ wrong principal amount, wrong coupon).
\item \textbf{1:} Multiple numerical inconsistencies; derived figures do not reconcile.
\item \textbf{2:} One numerical inconsistency that distorts interpretation, or unit-of-measure error.
\item \textbf{3:} Minor numerical slip (rounding, formatting); reviewer catches on routine read.
\item \textbf{4:} All numbers correct against source; derived figures reconcile.
\item \textbf{5:} All numbers correct, derived figures reconcile, and method of derivation is auditable.
\end{itemize}

\subsection*{D4. Workflow completeness}
\begin{itemize}[leftmargin=*,itemsep=0.1em]
\item \textbf{0:} Major required step skipped (e.g.\ omitted covenant block in extraction; missing risk section in profile).
\item \textbf{1:} Multiple steps skipped or partially completed.
\item \textbf{2:} One required step elided or thinly addressed.
\item \textbf{3:} All steps present; one is shallow.
\item \textbf{4:} All steps present and substantively addressed.
\item \textbf{5:} All steps present, substantively addressed, with appropriate depth and treatment of edge cases.
\end{itemize}

\subsection*{D5. Source discipline}
\begin{itemize}[leftmargin=*,itemsep=0.1em]
\item \textbf{0:} Fabricated content; assertions presented as fact that have no source basis.
\item \textbf{1:} Multiple instances of unsupported assumption or padding.
\item \textbf{2:} Notable padding or unsupported inferences distort the output.
\item \textbf{3:} Mostly disciplined; minor padding or weak inference visible.
\item \textbf{4:} Disciplined throughout; output stays within source.
\item \textbf{5:} Disciplined; explicit acknowledgement of evidence gaps where relevant.
\end{itemize}

\subsection*{D6. Decision usefulness}
\begin{itemize}[leftmargin=*,itemsep=0.1em]
\item \textbf{0:} Output cannot be used to advance the workflow.
\item \textbf{1:} Output requires substantial rework before use.
\item \textbf{2:} Output usable only with significant reviewer additions.
\item \textbf{3:} Output usable with reviewer judgement layered on.
\item \textbf{4:} Output advances the workflow; reviewer adds judgement, not content.
\item \textbf{5:} Output is directly usable; reviewer's role is verification rather than authorship.
\end{itemize}

\subsection*{D7. Reviewability and auditability}
\begin{itemize}[leftmargin=*,itemsep=0.1em]
\item \textbf{0:} Output's provenance is opaque; reviewer cannot reproduce.
\item \textbf{1:} Provenance partially recoverable; significant reconstruction effort required.
\item \textbf{2:} Provenance recoverable for some claims, not others.
\item \textbf{3:} Provenance recoverable with effort; not yet production-grade.
\item \textbf{4:} Provenance recoverable for all substantive claims.
\item \textbf{5:} Provenance recoverable, machine-readable, and replayable; full audit trail available.
\end{itemize}

\section{Demonstration Prompts}
\label{app:prompts}

Demonstration prompts for all five workflows (W1 DCM extraction, W2 precedent retrieval, W3 issuer profile, W4 transaction-comparable reasoning, W5 ECM extraction) are released alongside this paper in the public repository \citep{cmlrs2026repo}. To keep the paper focused, full prompts are not reproduced here. Each prompt is approximately 200--400 words and can be lifted from the repository or appended as additional sub-sections of this appendix at the venue's request.

\section{Worked Scoring Example: W1 / Sonnet 4.6 / HCA 2024 Senior Notes}
\label{app:worked}

This appendix shows one filled-in scoring instance end-to-end: source passage, model output, and rubric application. The cell is W1 (DCM transaction-terms extraction), model Claude Sonnet 4.6, document the HCA Inc.\ Rule~424(b)(5) prospectus supplement filed with the SEC on 9~August~2024 (US\$3.0bn senior notes across three tranches). The full document, the full model output, and the judge's full per-dimension JSON are in the repository at \texttt{eval/\allowbreak preprocessed/\allowbreak W1/}, \texttt{eval/\allowbreak outputs/\allowbreak W1\_sonnet.jsonl}, and \texttt{eval/\allowbreak scores/\allowbreak W1\_sonnet.jsonl} respectively, keyed on \texttt{cache\_key = "W1|\allowbreak claude-sonnet-4-6|\allowbreak 01\_hca\_2024-08-09|\allowbreak v1"}.

\subsection*{B.1 Source passage (excerpt)}

The W1 prompt asks the model to extract a structured term-sheet row from the prospectus supplement. The relevant cover-page fragment of the source reads (paraphrased to fit page width; the full source is in the repository):

\begin{quote}\small
\textit{HCA Inc., \$3,000,000,000 Senior Notes, consisting of: \$750,000,000 5.450\% Senior Notes due 2031; \$1,250,000,000 5.450\% Senior Notes due 2034; \$1,000,000,000 5.950\% Senior Notes due 2054. The new 2031 notes will be issued as additional notes under the indenture governing the \$1,000,000,000 aggregate principal amount of 5.450\% Senior Notes due 2031 issued on February 23, 2024 (existing 2031 notes); upon issuance, the aggregate principal amount of the 2031 notes will be \$1,750,000,000. Maturity dates: 2031 notes April 1, 2031; 2034 notes September 15, 2034; 2054 notes September 15, 2054. Public offering prices: 101.864\%, 99.854\%, 99.935\% respectively. The notes will be HCA Inc.'s senior obligations \dots and will be fully and unconditionally guaranteed on a senior unsecured basis by HCA Healthcare, Inc. \dots HCA Inc.\ intends to use the net proceeds of this offering for general corporate purposes, which may include the repayment of borrowings outstanding under its senior secured asset-based revolving credit facility.}
\end{quote}

\subsection*{B.2 Model output (excerpt)}

Sonnet 4.6 returned a structured Markdown table; one representative row is reproduced below.

\begin{quote}\small
\textbf{Principal amount:} Three tranches totalling \$3{,}000{,}000{,}000: (1) \$750{,}000{,}000 5.450\% Senior Notes due 2031 (new tap; aggregate 2031 series becomes \$1{,}750{,}000{,}000); (2) \$1{,}250{,}000{,}000 5.450\% Senior Notes due 2034; (3) \$1{,}000{,}000{,}000 5.950\% Senior Notes due 2054 -- \textit{Cover Page / ``The Offering -- Notes''}.

\textbf{Issue price:} (1) 2031 notes: 101.864\% of principal (plus accrued interest from February 23, 2024); (2) 2034 notes: 99.854\%; (3) 2054 notes: 99.935\% -- \textit{Cover Page pricing table}.

\textbf{Use of proceeds:} Net proceeds will be used for general corporate purposes, which may include repayment of borrowings under HCA Inc.'s senior secured ABL revolving credit facility -- \textit{Cover Page / ``Use of Proceeds -- S-15''}.

\textbf{Change-of-control provision:} If a Change of Control Triggering Event occurs, HCA Inc.\ is required to repurchase the notes of each series from holders on the terms described in the prospectus supplement; specific put price (typically 101\%) described in ``Description of the Notes -- Repurchase at the Option of Holders -- Change of Control Triggering Event'' (full detail truncated) -- \textit{Cover Page / ``Description of the Notes -- S-23''}.
\end{quote}

\subsection*{B.3 Rubric application}

The primary judge (Claude Sonnet 4.6 in independent judge mode, separate prompt) returned the per-dimension scores and one-sentence justifications shown in Table~\ref{tab:worked}. The aggregate is the equal-weighted mean across the seven dimensions: $(5+5+5+5+4+5+5)/7 = 4.86$.

\begin{table}[h]
\centering
\caption{Per-dimension scores and judge justifications for the worked example. Aggregate CM-LRS = 4.86.}
\label{tab:worked}
\small
\begin{tabularx}{\textwidth}{@{}lcX@{}}
\toprule
\textbf{Dimension} & \textbf{Score} & \textbf{Judge justification (verbatim)} \\
\midrule
D1 Factual accuracy & 5 & ``All stated facts -- issuer, guarantor, tranche sizes, coupons, maturities, issue prices, use of proceeds, and structural ranking -- are precisely correct against the source document with no errors or misstatements.'' \\
\addlinespace
D2 Evidence traceability & 5 & ``Every field is tagged with a specific section reference (e.g., `Cover Page pricing table,' `The Offering -- Maturity Date,' `Description of the Notes -- S-23'), enabling direct verification against the source.'' \\
\addlinespace
D3 Numerical consistency & 5 & ``All figures are internally consistent and match the source: \$750M + \$1{,}250M + \$1{,}000M = \$3{,}000M total; aggregate 2031 series correctly stated as \$1{,}750M; issue prices 101.864\%, 99.854\%, 99.935\% match the pricing table exactly.'' \\
\addlinespace
D4 Workflow completeness & 5 & ``All 15 requested fields are addressed, with truncation-limited fields explicitly marked `NOT STATED in the excerpted text' rather than omitted, ensuring full coverage of the extraction template.'' \\
\addlinespace
D5 Source discipline & 4 & ``The model correctly marks unavailable details as `NOT STATED' and avoids fabrication, though it adds a parenthetical inference (`typically 101\%' for CoC put price and `typically New York law') that goes slightly beyond the truncated source text.'' \\
\addlinespace
D6 Decision usefulness & 5 & ``The structured table format with tranche-level breakdown, aggregate series size, accrued interest note, and explicit flags for truncated covenant details gives a banker or analyst an immediately actionable term-sheet summary.'' \\
\addlinespace
D7 Reviewability & 5 & ``Section references on every row, clear `NOT STATED' flags with explanatory context, and a clean tabular format allow a reviewer to rapidly cross-check each claim against the source document.'' \\
\bottomrule
\end{tabularx}
\end{table}

\subsection*{B.4 Reading the result}

The cell scores 4.86 overall: the output would clear a CM-LRS\,$\geq$\,4.0 deployment gate with a single-dimension dock on D5 (source discipline) for the parenthetical inferences. The dock is informative rather than disqualifying: a deployment-readiness review would either tighten the prompt to suppress hedging inferences or accept them as a known artefact of truncation. The example illustrates the rubric's intended use: the aggregate is the headline, but the dimension that drops is the one a reviewer should look at, and the judge's justification gives the reviewer the specific text span to check.

\end{document}